\documentclass[10pt,twocolumn,letterpaper]{article}

\usepackage{wacv}
\usepackage{times}
\usepackage{epsfig}
\usepackage{graphicx}
\usepackage{amsmath}
\usepackage{amssymb}
\usepackage{booktabs}
\usepackage{comment}

\definecolor{cvprblue}{rgb}{0.21,0.49,0.74}
\usepackage[pagebackref,breaklinks,colorlinks,citecolor=cvprblue]{hyperref}
\usepackage{enumitem}

\def\wacvPaperID{} 

\wacvapplicationstrack 

\wacvfinalcopy 

\begin{document}

\title{Vehicle Vectors and Traffic Patterns from Planet Imagery}

\author{Adam Van Etten\\
Planet\\
{\tt\small adam.van.etten@federal.planet.com}
}

\maketitle
\thispagestyle{empty}

\begin{abstract}
We explore methods to detect automobiles in Planet imagery and build a large scale vector field for moving objects. Planet operates two distinct constellations: high-resolution SkySat satellites as well as medium-resolution SuperDove satellites.  We show that both static and moving cars can be identified reliably in high-resolution SkySat imagery.  We are able to estimate the speed and heading of moving vehicles by leveraging the inter-band displacement (or ``rainbow'' effect) of moving objects.  Identifying cars and trucks in medium-resolution SuperDove imagery is far more difficult, though a similar rainbow effect is observed in these satellites and enables moving vehicles to be detected and vectorized.  
The frequent revisit of Planet satellites enables the categorization of automobile and truck activity patterns over broad areas of interest and lengthy timeframes.

\end{abstract}

\section{Introduction}
\label{sec:intro}

Cars and trucks are primary components of the global transportation sector.
The close link between vehicular traffic and human activity means that traffic patterns are a useful proxy for important metrics such as  social unrest, instability, or economic productivity.  

The near daily coverage of the Earth's landmass with the Planet satellite constellation begs an interesting question: can vehicular activity be analyzed on a daily cadence from satellite imagery?  In this paper demonstrate that Planet's dual constellations each provide unique insights to this question.

We rely upon Planet satellite imagery for our analysis.  Planet operates two primary constellations.  The PlanetScope constellation consists of hundreds of small 3U SuperDove satellites. Each SuperDove is a 3U CubeSat, where U stands for 10 x 10 x 10 cm stowed dimensions.  These satellites provide nearly daily coverage of the Earth's landmass, and provide imagery at 3 meter ground sample distance (GSD), meaning that one imaging pixel is $3 \times 3$ m in extent.
A higher resolution constellation consists of $\sim20$ far larger SkySat satellites that provide tasked 0.50 meter imagery over smaller areas.  

We use this Planet imagery to study the transportation sector,  specifically vehicular patterns.  Establishing vehicle counts over time is a well explored computer vision analytic for the small areas of interest (AOIs) covered by tasked satellites.  On the other hand, Inferring vehicle speed and heading is a poorly studied phenomenon,  particularly on a global scale.  To our knowledge, prior to this work no models exist to algorithmically create vehicle vector fields with Planet \cite{planet} satellite imagery.  This ability to dive deeper than simple counts and explore vehicle speed and heading opens up further analytics possibilities, which we explore in later sections.  



\section{Prior Work}
\label{sec:prior}

A number of prior works have explored how to detect moving vehicles in satellite imagery.  
Over a decade ago, \cite{kraub} showed that moving vehicles could be identified in high-resolution single satellite images by finding the difference between object centroids in various bands.
\cite{drouyer} used OpenStreetMap \cite{openstreetmap} road masks, image differencing, and top-hat morphological filters to identify moving vehicles in sequences of SuperDove (3m resolution) Planet images.  This technique assumed that ``vehicles are lighter than the road at least in one RGB channel. The detector will therefore detect white, silver, blue, red, brown or green vehicles, but won’t detect black or grey vehicles.''

\cite{rs13020208} explored spatial temporal traffic patterns of vehicles with Planet data during the initial stages of the COVID-19 pandemic.  These authors relied on simple  (yet effective) morphology and intensity manipulations (similar to \cite{drouyer}) of Planet imagery to identity moving vehicles, and requires the use of OpenStreetMap to identify possible locations of vehicles.  These authors report an F1 score of 0.68 when assigning a true positive detection as a detection with {\it any} overlap with ground truth (intersection over union (IOU) $ > 0.0$). The authors restricted detections to roads by masking out regions non-coincident with OSM roads.

Most recently, \cite{keto} showed that the velocity of aircraft can be manually deduced from PlanetScope imagery using the difference in location between sequential observations of PlanetScope imagery.  \cite{synth} showed that high altitude balloons could be located in PlanetScope imagery based on the telltale signature of offset red, green, and blue circles.

These prior works are impressive in their own right and help inspire our work, particularly \cite{rs13020208} and \cite{keto}.  Where our method differers is in the ability to run automatically at scale for both high resolution (SkySat) and medium resolution (PlanetScope) imagery without any external data sources required (e.g. OpenStreetMap).  Our use of high-resolution SkySat imagery in conjunction with PlanetScope imagery also enables a more extensive and precise ground truth dataset.
Furthermore, to our knowledge no prior works have extracted a large scale vector field of moving vehicles over multi-month time periods.




\section{Dataset}
\label{sec:dataset}

\begin{figure}
\begin{tabular}{l}
  \includegraphics[width=0.99\linewidth]{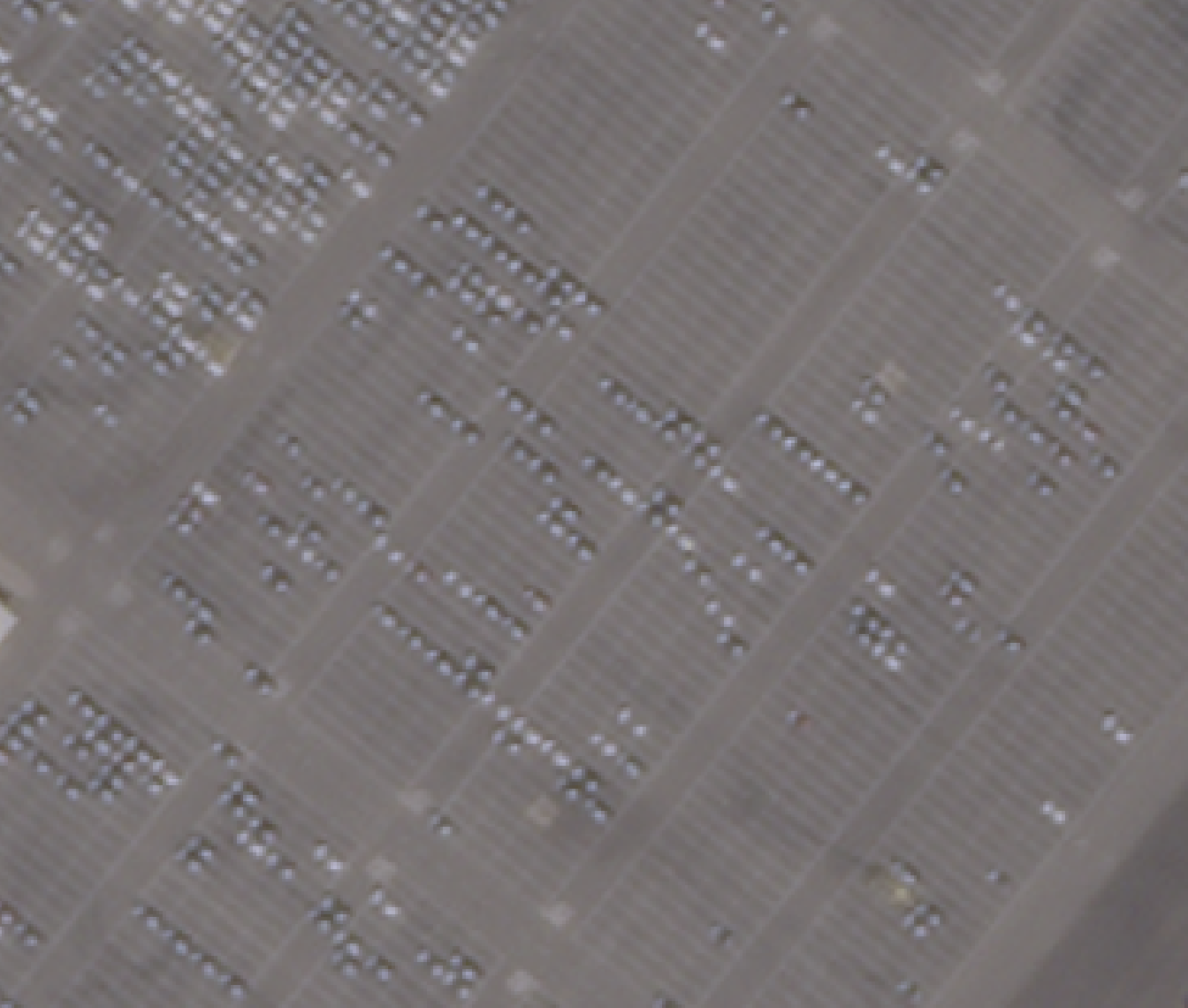} \\  
  \includegraphics[width=0.99\linewidth]{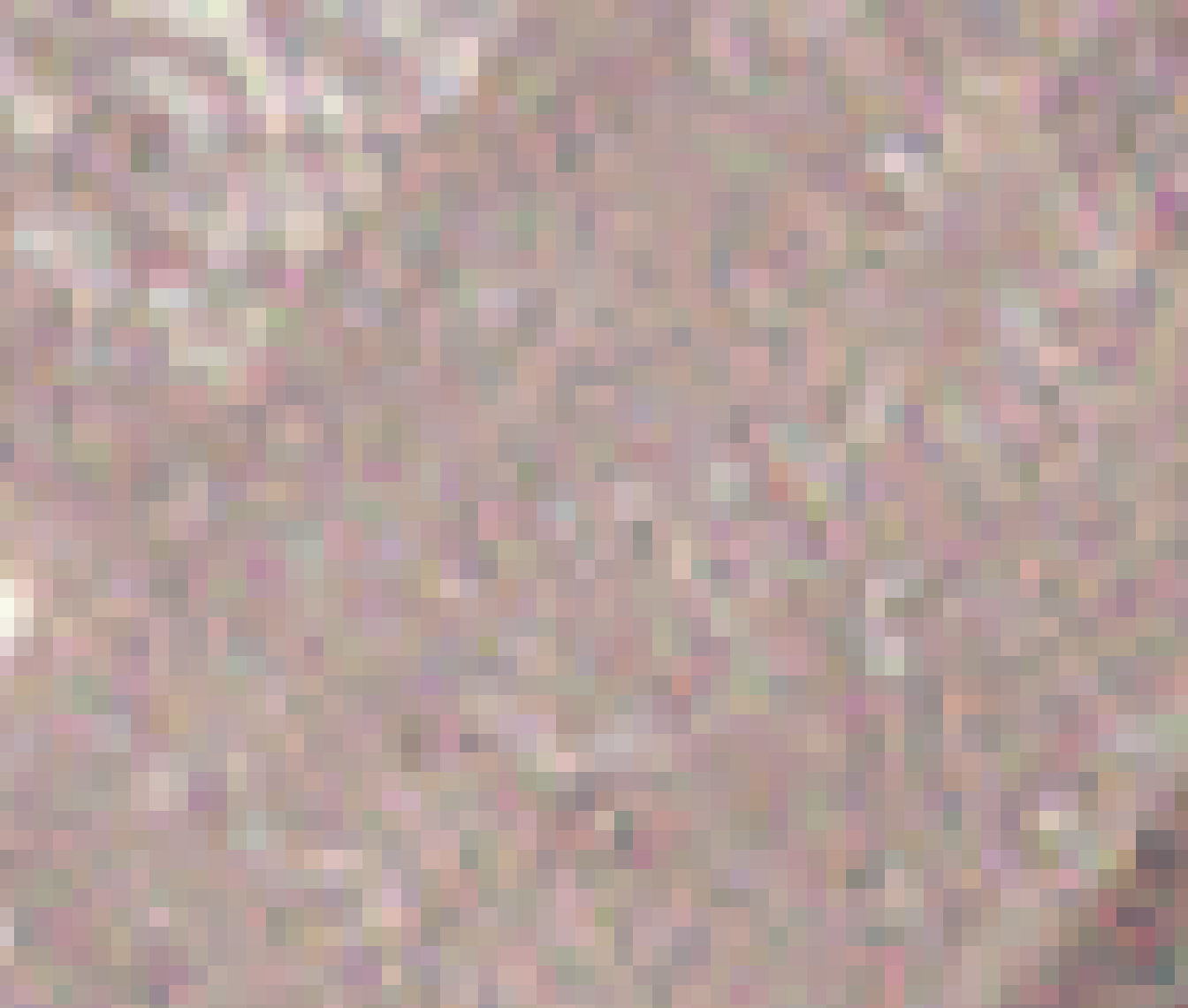} \\
  [-0pt] 
\end{tabular}
\caption{Concurrent, co-located SkySat (top) and PlanetScope (bottom) imagery of a parking lot in Long Beach, CA. }
\label{fig:f1_plots}
\end{figure}

Identifying and labeling individual cars is nigh impossible in PlanetScope imagery, though feasible for SkySat imagery, see Figure \ref{fig:f1_plots}.  We show in Section \ref{sec:concurrent_data}, however, that the higher resolution SkySat constellation can be highly complementary to PlanetScope for locating small vehicles.

\begin{figure}
\begin{tabular}{l}
  \includegraphics[width=0.98\linewidth]{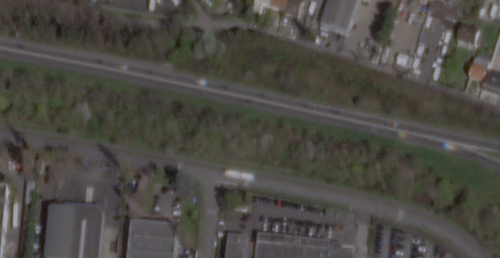} \\
  \includegraphics[width=0.98\linewidth]{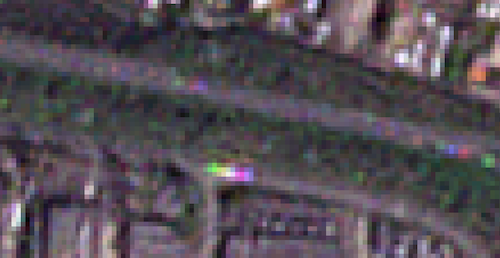} \\  
\end{tabular}
\caption{Concurrent SkySat (top) and PlanetScope (bottom) imagery of moving vehicles. Note the rainbow effect of moving vehicles in both images.}
\label{fig:f2_plots}
\end{figure}

Both constellations use a sequential sensor, where the same portion of the Earth is imaged sequentially by each imaging band as the satellite passes overhead.  This time delay between imaging bands produces a ``rainbow'' effect for moving objects.  This phenomenon has been well documented in aircraft \cite{keto} or balloons \cite{synth}, though less studied for terrestrial vehicles.  Yet terrestrial vehicles often have a high enough velocity (especially on highways) to demonstrate significant rainbows in both SkySat and PlanetScope, see Figure \ref{fig:f2_plots}.

The time delta between the collection of the green and blue bands in SuperDove is $\sim800$ ms (there are significant caveats to this $\sim800$ ms value that we will discuss in Section \ref{sec:veh_vec}).  With 3 meter pixels, this implies that for a vehicle velocity of 15 m/s ($\sim55$ km/h) we would see a $\sim4$-pixel shift between green and blue.  Therefore for objects traveling at least 50 km/h, we expect to see some sort of motion blur or “rainbow,” with the shift becoming much more obvious as speed increases.  At highway speeds of  90 km/h the “rainbow” effect is much more pronounced.  Therefore, highways are a good target for potentially locating moving vehicles.  Moving cars are difficult (though not impossible) to see, though trucks and buses return a strong signal see Figure \ref{fig:f2_plots}.


\subsection{SkySat Labeled Dataset}
\label{sec:skysat_data}

We hand-label a small dataset of SkySat images. We use the 3-band RGB (red-green-blue) ``Visual'' SkySat data \cite{planet_api} product since the motion “rainbow” shows up prominently in this imagery. There are 2,246 total cars in the dataset with 1,705 static cars and 571 moving cars labeled, and a 75/25\% train/test spilt.

\begin{figure}
  \begin{center}
    \includegraphics[width=0.99\linewidth]{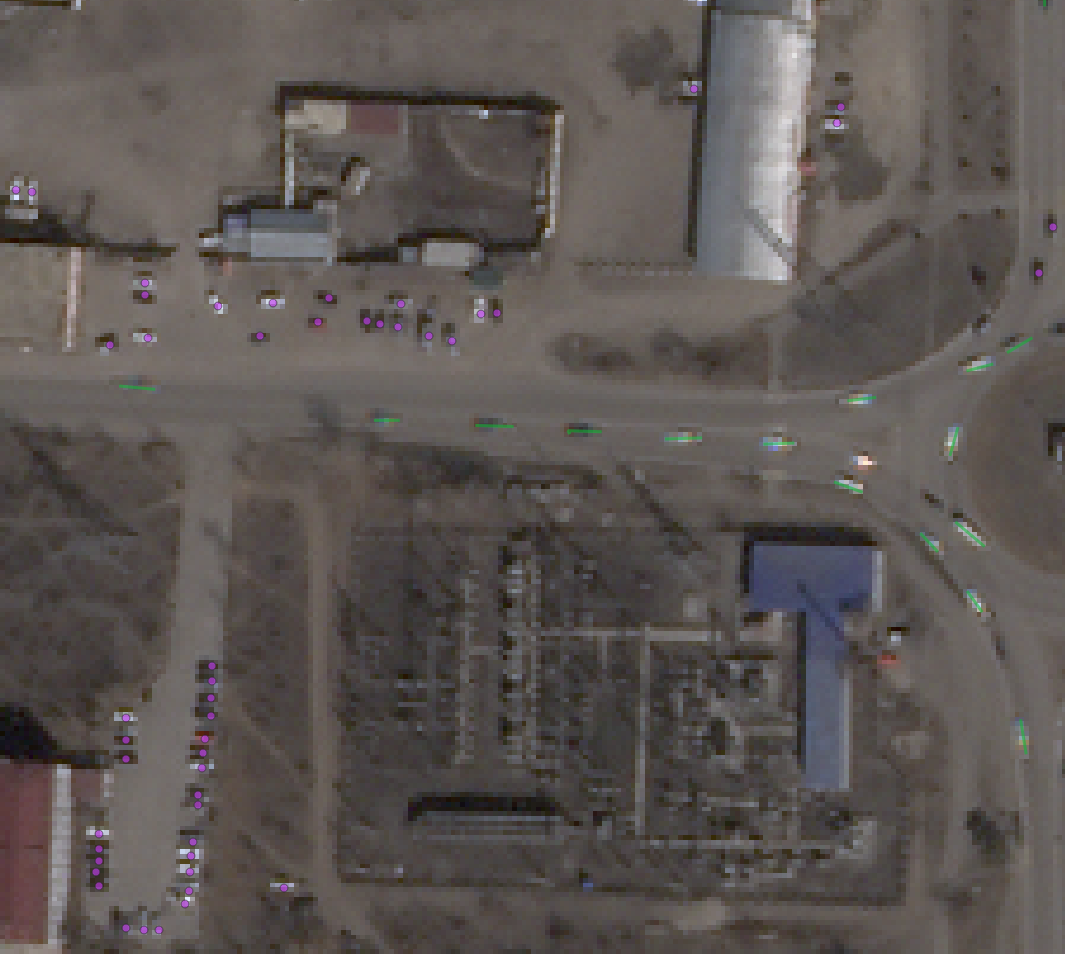}
  \end{center}
  \vspace{-0pt}
  \caption{SkySat car labeling schema. Static cars are labeled with a magenta dot, with moving cars labeled with green linestring}
  \label{fig:train_data0}
  \vspace{-8pt}
\end{figure}

Static cars are largely the same size and shape when viewed from overhead, so we adopt the schema of \cite{cowc} and simply assign a point label to the centroid of each car.  For moving vehicles we adopt a simple linestring label along the vehicle rainbow, see Figure \ref{fig:train_data0}.

\subsection{SkySat / PlanetScope Concurrent Dataset}
\label{sec:concurrent_data}

Labeling cars in PlanetScope imagery is a far more difficult than with SkySat.  As evidenced by Figure \ref{fig:f1_plots}, localizing individual static cars is improbable with the 3m resolution of PlanetScope.  Yet Figure \ref{fig:f2_plots} illustrates that motion induced rainbows may be detectable.  

\begin{figure}
\begin{tabular}{ll}
 \includegraphics[width=0.46\linewidth]{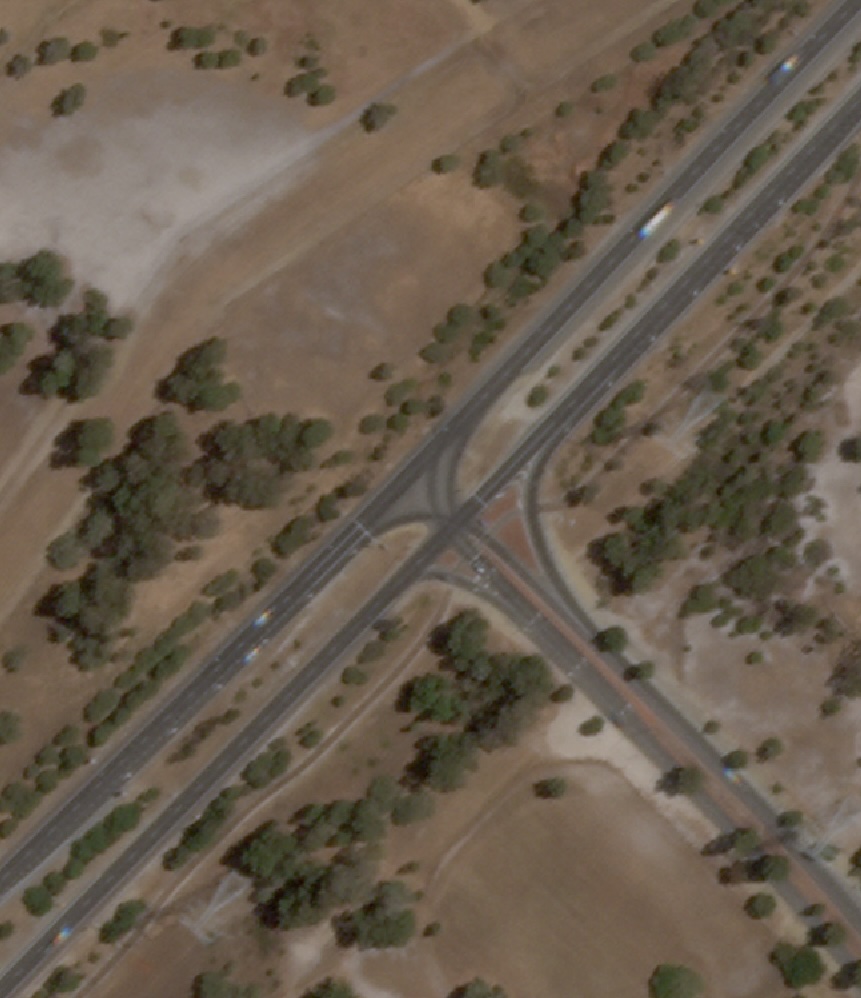} &  \includegraphics[width=0.46\linewidth]{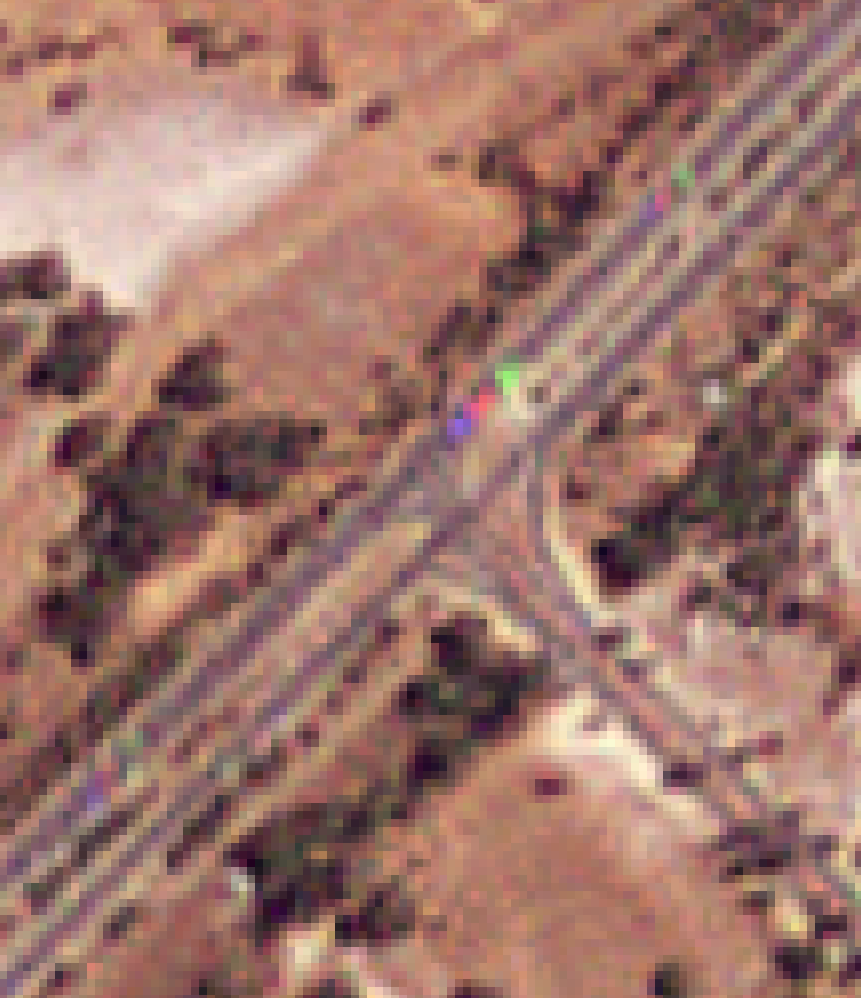} \\  
 \includegraphics[width=0.46\linewidth]{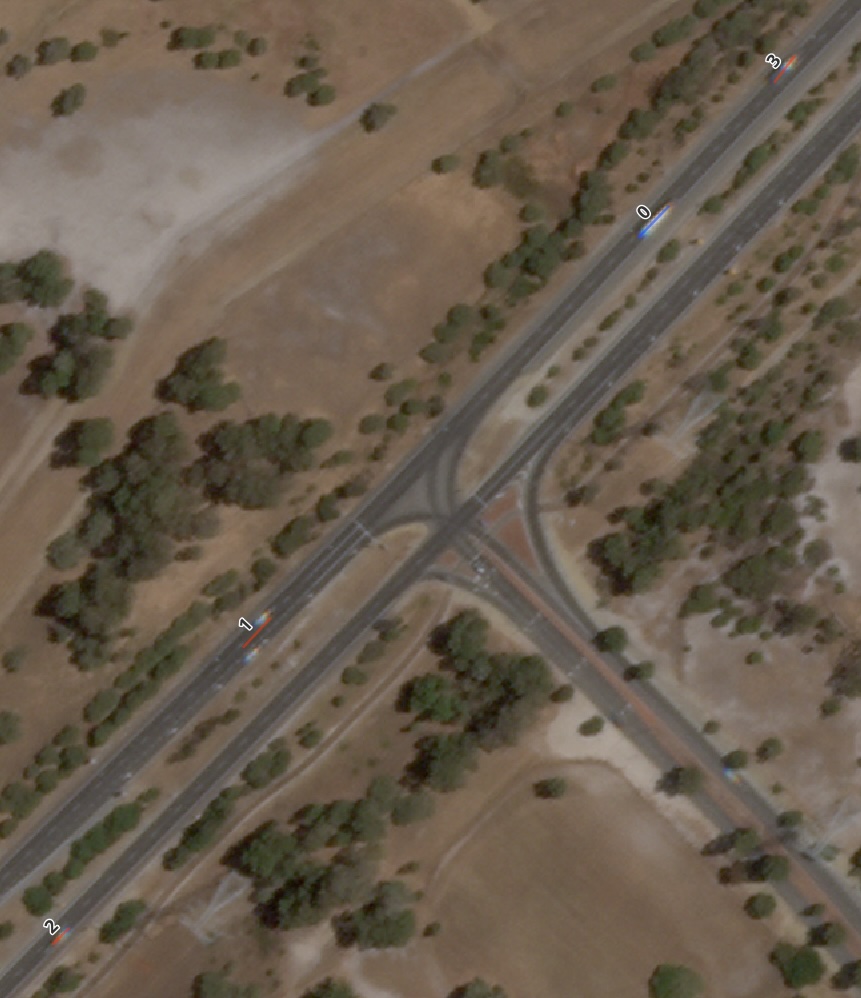} & \includegraphics[width=0.46\linewidth]{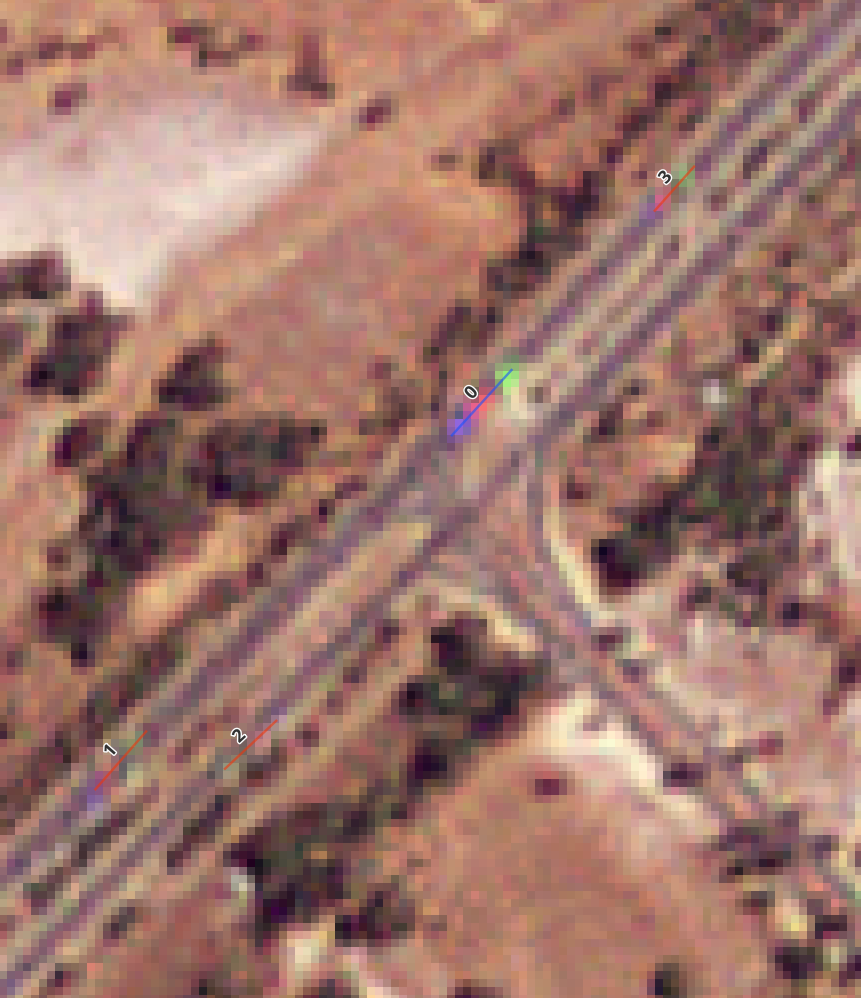} \\  
\end{tabular}
\caption{Concurrent SkySat (left) and PlanetScope (right) imagery of moving vehicles taken over Perth, Australia with $\Delta t < 1 \, {\rm s}$.
	{\bf Top Left:} Raw SkySat image. 
	{\bf Top Right:} Raw PlanetScope image. 
	{\bf Bottom Left:} SkySat image with labeled vehicles overlaid. ``0'' is a large vehicle (bus or truck), whereas ``1'', ``2'', ``3'' correspond to moving cars. ``1'' actually constitutes two unique cars, though we combine these into one label since these will be indistinguishable in PlanetScope; such labels are not used for SkySat training.
	{\bf Bottom Right:} PlanetScope image with labeled vehicles overlaid. 
	 }
\label{fig:concurrent}
\end{figure}

In order to confidently label moving vehicles in PlanetScope we leverage a unique feature of the dual Planet constellations: concurrent collection (or crossover) events.  
Accordingly, we gather imagery over 10 populated areas, with a collection delta between SkySat and PlanetScope systems of $0.05 \leq \Delta t \leq 3.0$ seconds. Even at these low $\Delta t$ values, vehicles move significantly between collections.  At highway speeds of $80 \, \rm{km/h} \approx 20 \,\rm{m/s}$, vehicles will be displaced by $1 - 60 {\rm m}$ for our $\Delta t$ values.  Therefore, unfortunately one cannot simply label vehicles in SkySat and apply those labels to the lower resolution PlanetScope imagery.  What is possible, however, is to use SkySat imagery to aid manual labeling of moving vehicles in PlanetScope imagery.  This is accomplished by switching back and forth between PlanetScope and SkySat imagery (we use QGIS \cite{QGIS}), and visually extrapolating which SkySat vehicles correspond to the rainbows observed in PlanetScope.  This helps reduce false positives caused by spurious rainbows that rarely occur in PlanetScope, and also helps differentiate between moving cars and moving trucks/buses.  See Figure \ref{fig:concurrent} for labeling specifics.

\subsection{PlanetScope Labeled Dataset}
\label{sec:ps_data}

Using the concurrent imagery detailed in Section \ref{sec:concurrent_data} and the comparative labeling process illustrated in Figure \ref{fig:concurrent}, we build a dataset of moving cars and trucks in PlanetScope, assigning a linestring label to each observed (and confirmed via SkySat) vehicle rainbow.  Labeling in this manner is nontrivial, and our final training dataset is rather small: 546 cars and 151 trucks (697 total).  

\section{Detection Algorithm}
\label{sec:algo}

We implement a pixel segmentation + polygon inference process to detect and classify vehicles.  For this problem, a segmentation algorithm is preferable to a bounding box object detection framework (e.g. YOLT \cite{yolt}) since objects very elongated in shape, and vehicles are frequently very densely packed.  Bounding box detectors such as YOLT perform best with objects that are well spread out and have an aspect ratio (length / width) $\lesssim 2$.  

\subsection{Segmentation}
\label{sec:segmentation}

\begin{figure}
  \begin{center}
    \includegraphics[width=0.99\linewidth]{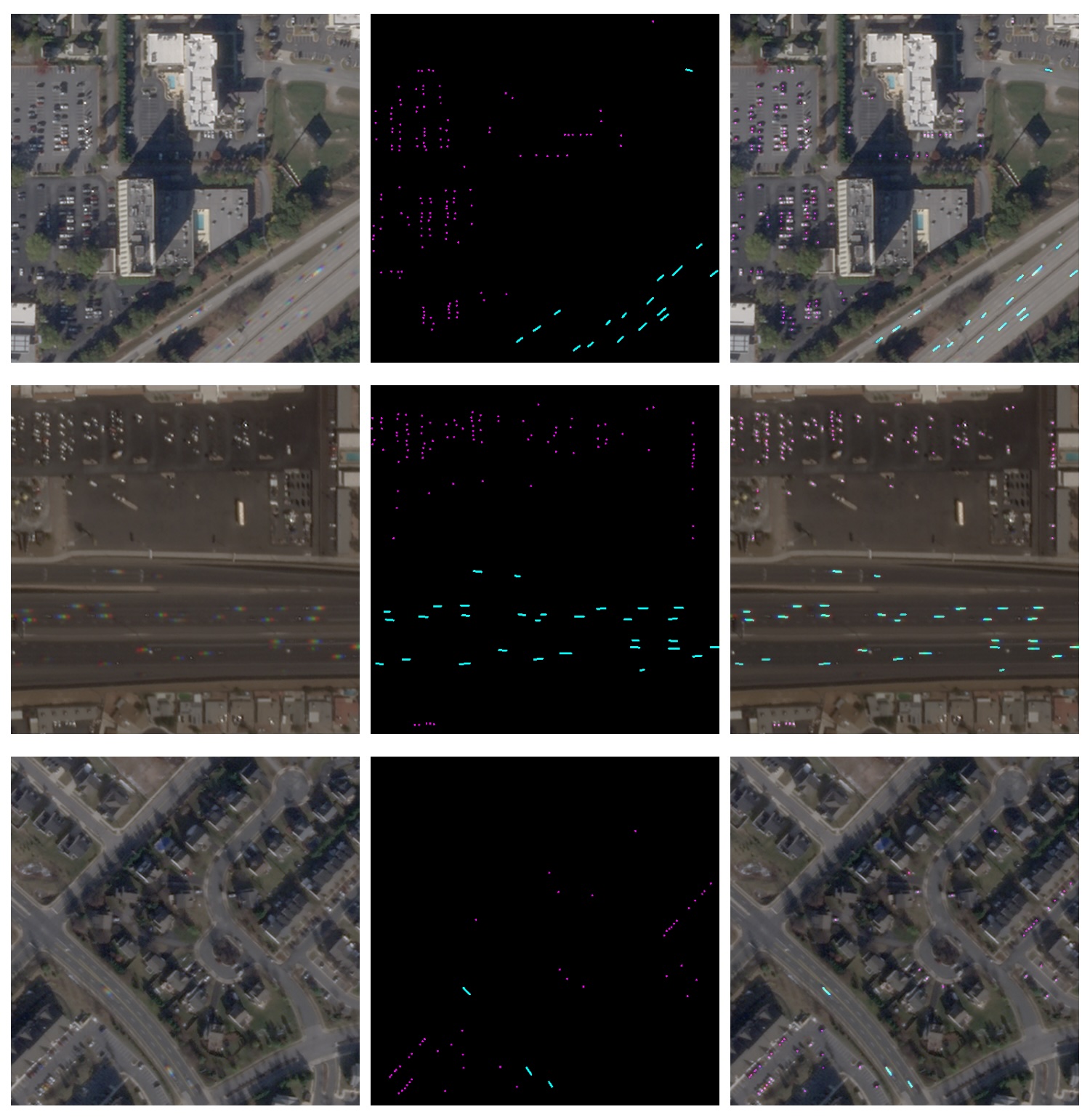}
  \end{center}
  \vspace{-12pt}
  \caption{SkySat images and training masks for car detection. {\bf Left:} Training image. {\bf Middle:} Multi-class training mask with a 0.75 meter buffer around labels; magenta dots are static cars, and cyan lines are moving vehicles. {\bf Right:} Training mask overlaid on the training image.}
  \label{fig:ss_masks}
\end{figure}

\begin{figure}
  \begin{center}
    \includegraphics[width=0.99\linewidth]{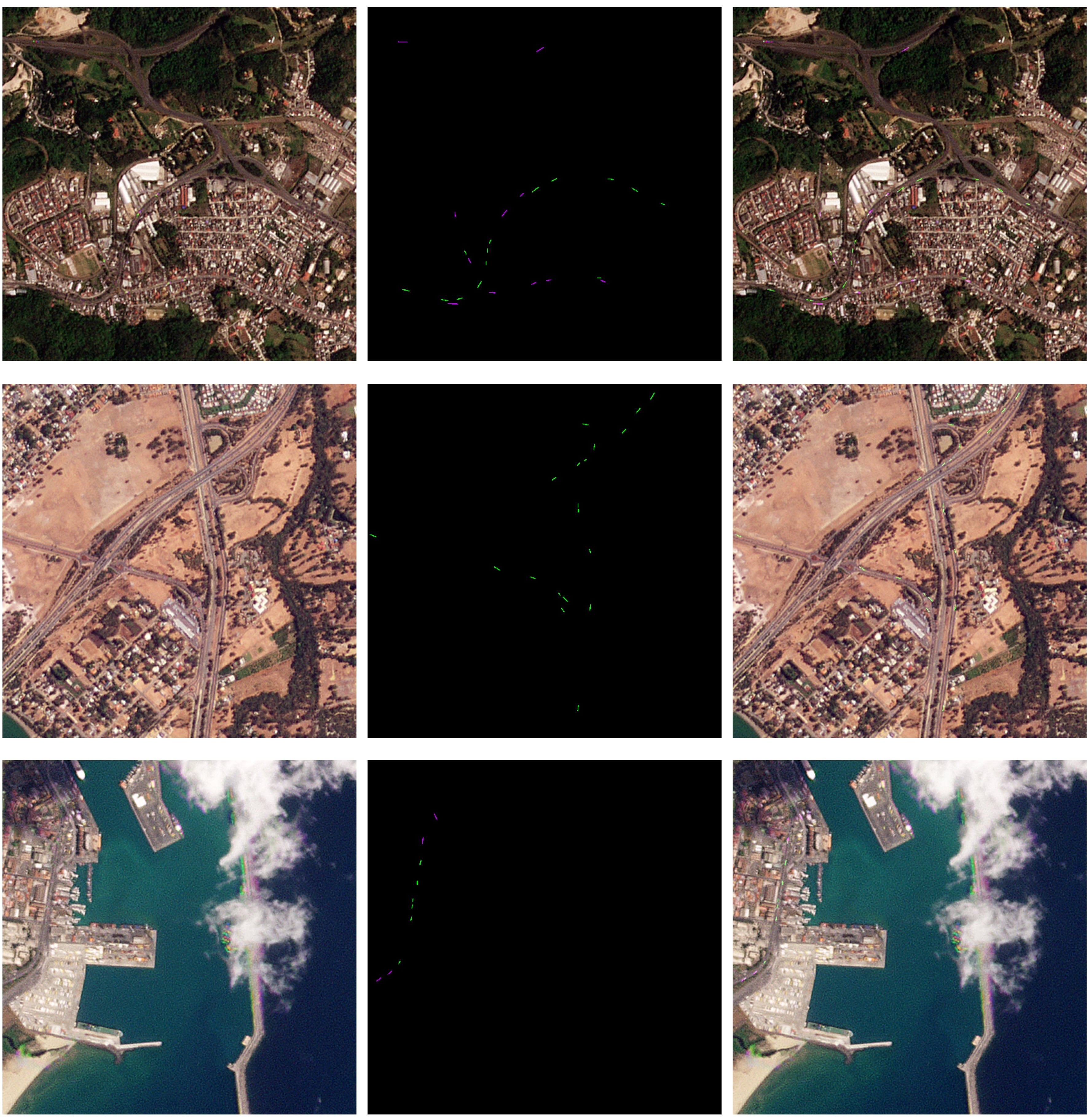}
  \end{center}
  \vspace{-4pt}
  \caption{PlanetScope images and training masks for moving vehicle detection. {\bf Left:} Training image. {\bf Middle:} Multi-class training mask with a 2 meter buffer around linestring labels; green denotes cars, and purple denotes trucks/buses. {\bf Right:} Training mask overlaid on the training image. Note the cloud-induced rainbows in the bottom row.}
  \label{fig:ps_masks}
\end{figure}

For segmentation we use the open source CRESI \cite{cresi} deep learning segmentation codebase. We render our linestring and point labels into 2-channel masks by creating a buffer around each vehicle.  This buffer is set to 0.75 m in SkySat, and 2 m in PlanetScope imagery, see Figure \ref{fig:ss_masks}, \ref{fig:ps_masks}. For SkySat, Layer0 of the mask corresponds to static cars, and Layer1 denotes moving cars. For PlanetScope, Layer0 of the mask corresponds to moving cars, and Layer1 denotes moving trucks/buses. Using these imagery+mask datasets, we train ResNet34 \cite{resnet} + UNet \cite{unet} segmentation models for 50 epochs.

\subsection{Vehicle Localization and Vector Inference}
\label{sec:veh_vec}

\begin{figure}
  \begin{center}
    \includegraphics[width=0.99\linewidth]{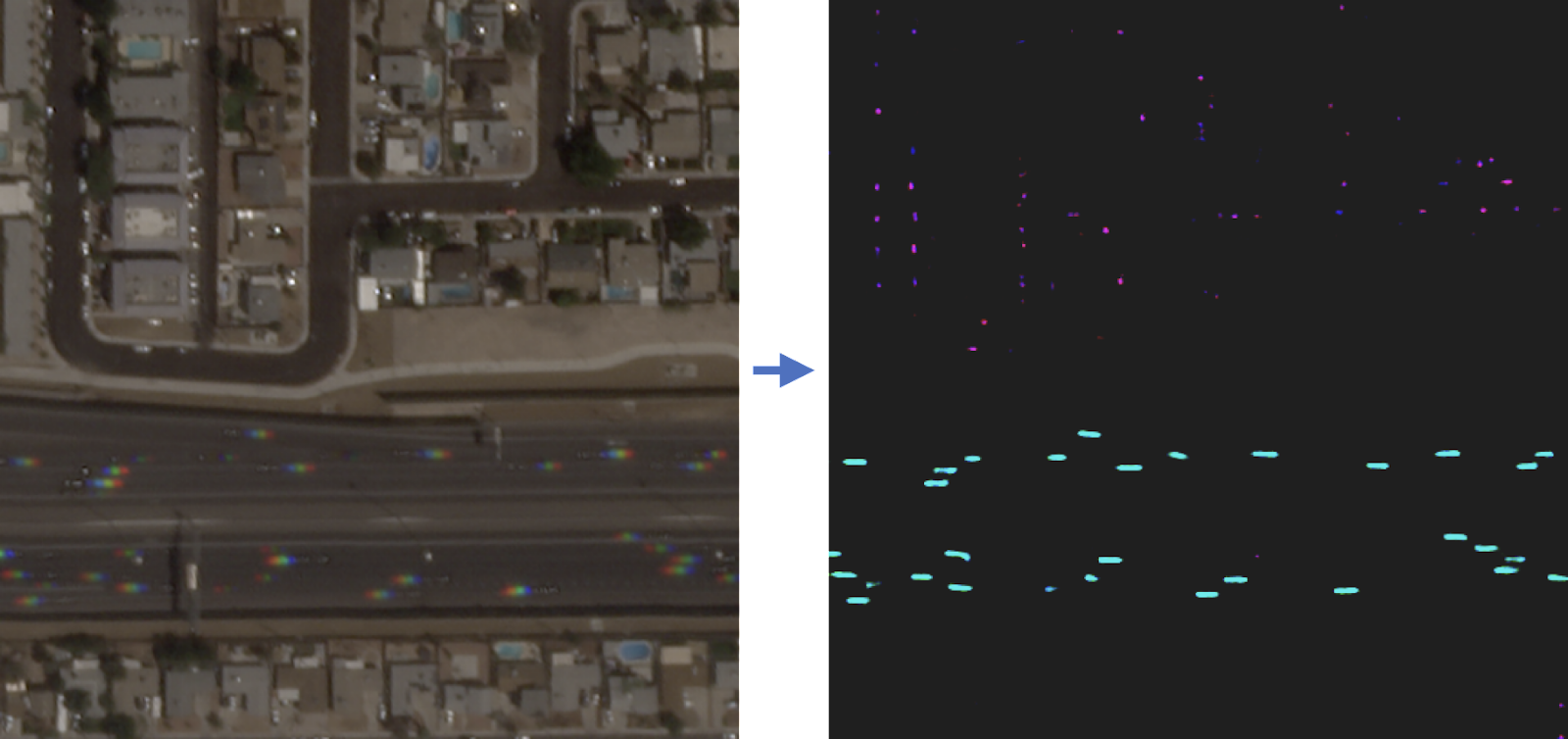}
  \end{center}
  \vspace{-0pt}
  \caption{{\bf Left:} SkySat test imagery. {\bf Right:} Predicted mask from the SkySat segmentation model, showing static cars in magenta and moving cars in cyan.}
  \label{fig:preds0}
\end{figure}

\begin{figure}
  \begin{center}
    \includegraphics[width=0.99\linewidth]{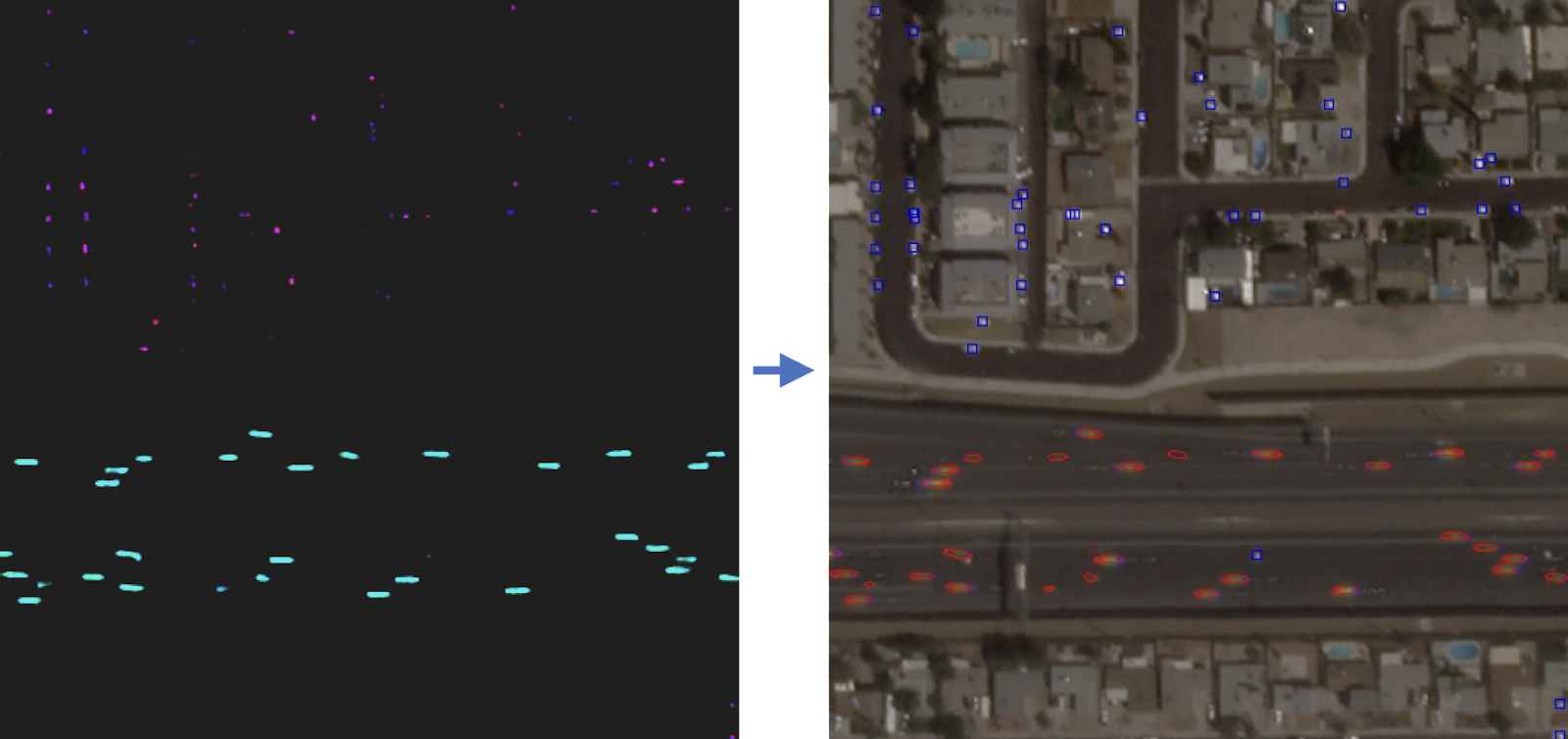}
  \end{center}
  \vspace{-0pt}
  \caption{{\bf Left:} Inference mask for the test image in Figure \ref{fig:preds0}. {\bf Right:} Inferred static cars (blue boxes) and moving cars (red ellipses) overlaid on the test imagery.}
  \label{fig:preds1}
\end{figure}

Applying our trained SkySat segmentation model to the test set yields predictions akin to Figure \ref{fig:preds0}. The raw prediction mask is post-processed with the following steps:


\begin{enumerate}[leftmargin=2\parindent]
    \item Threshold the prediction mask
    \item Extract contours of the thresholded mask
    \item Extract vehicle polygons (see Figure \ref{fig:preds1})
        \begin{enumerate}
            \item Static: Create a $3 \times 3$ m square at contour center.
            \item Moving: Fit an ellipse to each contour.
    	\end{enumerate}
    \item Extract heading (direction of motion)
    \item Infer magnitude (speed)
\end{enumerate} 

\begin{figure}
  \begin{center}
    \includegraphics[width=0.99\linewidth]{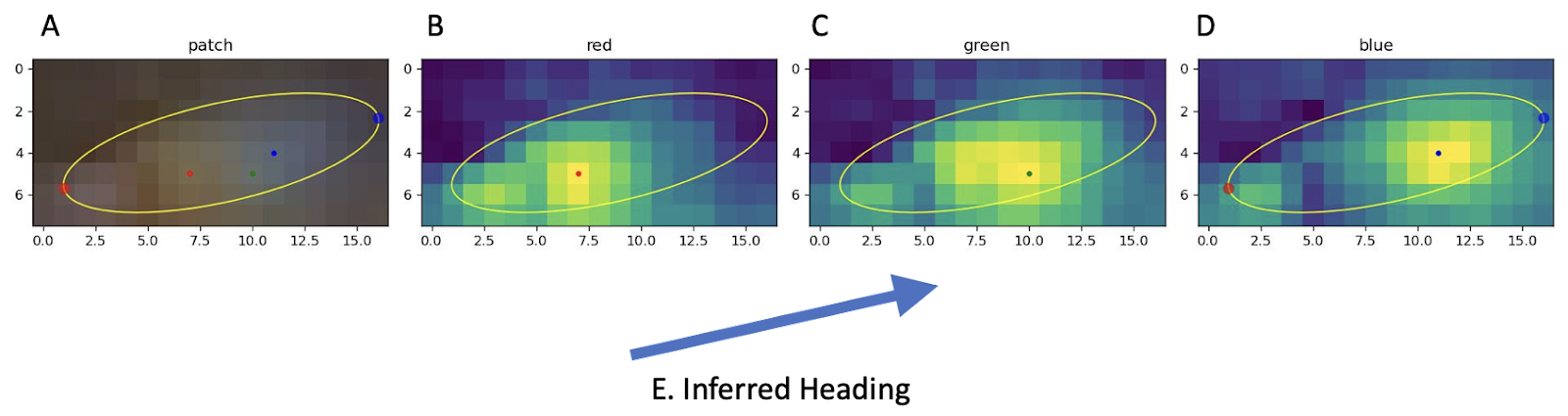}
  \end{center}
  \vspace{-0pt}
  \caption{Vehicle heading determination. {\bf A}: Image patch with fitted ellipse overlaid in yellow. {\bf B, C, D}: Red, green and blue channels (respectively) for the image patch, with the peak in each band plotted as a dot. Note that the red and blue centroids are displaced. We use the difference in red and blue peaks to determine the inferred heading ({\bf E}). In this example, the inferred heading is 81 degrees, with magnitude 54 km/h.}
  \label{fig:vec0}
\end{figure}

For moving vehicles, once we fit an ellipse to each contour detection (Step 3b), we subsequently leverage the vehicle rainbow to determine velocity.  The major axis of the fitted ellipse gives the heading of the car, albeit with a 180$^{\circ}$ ambiguity. To resolve this ambiguity, we extract the imagery pixels around each fitted ellipse (Figure \ref{fig:vec0}A). The individual red, green, and blue channels are extracted from this chip, and the location of the peak value of each band is recorded (the dots in Figure \ref{fig:vec0}B, \ref{fig:vec0}C, \ref{fig:vec0}D). We use the difference in locations of the red and blue peaks to resolve the direction of the red-to-blue vector and infer vehicle heading.  

The length of the major axis of the ellipse is proportional to the speed of the object. (In reality, the translation of motion blur length to object speed will depend on a number of factors such as nadir and azimuth angle, but these are secondary effects that we will tackle in future refinements.)
For the SkySat RGB Visual product, we assume the time delta between red and blue bands is 
$\approx 560$ ms.  An important feature of the imagery is that the Visual product is a composite of many scenes, and the precise scenes used for the composite are not recorded.  Therefore the exact temporal spacing between bands will vary significantly from scene to scene.  We therefore assign very conservative $30\%$ errors to the speed estimate.
Therefore, with 0.5m pixels, speed $|v_{ss}|$ scales with rainbow length in pixels ($d_{pix}$) as:
\begin{equation}
	|v_{ss}| = 0.5 \, d_{pix} / 0.560 {\rm \, m/s}  = 3.2 \pm 1.0 \, d_{pix} {\rm \, km/h} 
\end{equation}
For the PlanetScope SuperDove satellites, the time delta between red and blue bands is $\approx 800$ ms.  As with SkySat imagery,  the precise scenes used for the composite are not recorded and we assign very conservative $30\%$ errors to the speed estimate. Therefore, with 3m pixels, speed $|v_{ps}|$ scales with rainbow length in pixels ($d_{pix}$) as:
\begin{equation}
	|v_{ps}| = 3 \, d_{pix} / 0.800 {\rm \, m/s}  = 13 \pm 4 \, d_{pix} {\rm \, km/h} 
\end{equation}

This allows us to estimate the vector field of moving vehicles in both SkySat and PlanetScope imagery, which we detail further in the next section.  

Algorithmic runtimes for SkySat are $\approx 60 \, {\rm km^2/min/GPU}$.  Runtimes for PlanetScope are 
significantly faster,
at $\approx 600 \, {\rm km^2/min/GPU}$.


\section{Results}
\label{sec:results}

\subsection{SkySat Results}

\begin{figure}
  \begin{center}
    \includegraphics[width=0.98\linewidth]{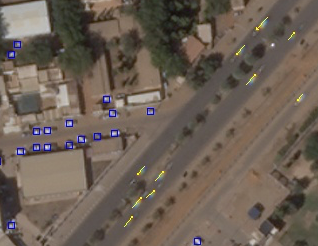}
  \end{center}
  \vspace{-8pt}
  \caption{Inferred velocity vectors in a SkySat test image. Blue denotes static cars, while the length of the yellow vectors are proportional to vehicle speed.}
  \label{fig:vec_field}
\end{figure}

Combining the inferred vehicle mask with the post processing of Section \ref{sec:veh_vec}, we are able to create a vector field for moving vehicles, as shown  in Figure \ref{fig:vec_field}.  Table \ref{tab:skysat_results} displays quantitative results of F1 scores as well as the aggregate count fraction.  Aggregate count fraction ($count_{frac}$) is simply the number of predicted vehicles $N_{pred}$) divided by the number of ground truth vehicles ($N_{gt}$).  A count fraction of 1.0 is desired.
\begin{equation}
	count_{frac} = N_{pred} / N_{gt}
	\label{eqn:count_frac}
\end{equation}

Since we are localizing small objects, we assign a true positive as any prediction with $IOU \geq 0.25$ to ground truth. Given our small dataset size, our F1 scores are poor, though aggregate counts are within $3\%$ of ground truth.

\begin{table}
  \centering
  \begin{tabular}{@{}lcc@{}}
    \toprule
     {\bf Class} & {\bf F1} & {\bf $count_{frac}$} \\
    \midrule
    {Static} & $0.39$ & $1.03$ \\
    {Moving} & $0.81$  & $1.02$\\
    \midrule
    {Mean} & $0.60$  & $1.03$\\    
    \bottomrule
  \end{tabular}
  \vspace{2pt}
  \caption{SkySat Performance.}
  \label{tab:skysat_results}
  \end{table}

\subsection{PlanetScope Results}

\begin{figure}
  \begin{center}
  \begin{tabular}{ll}
 \includegraphics[width=0.97\linewidth]{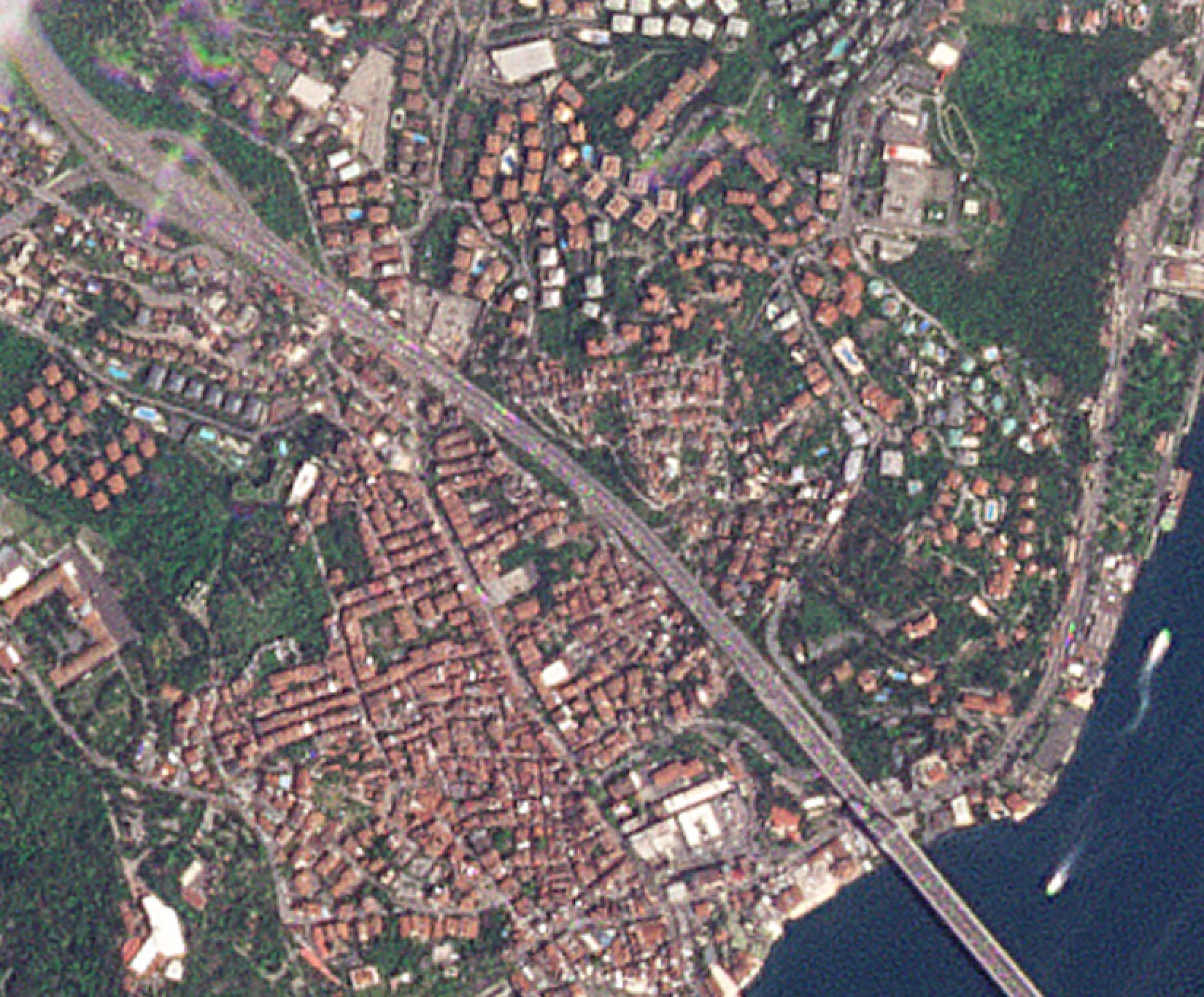} & \\
 \includegraphics[width=0.97\linewidth]{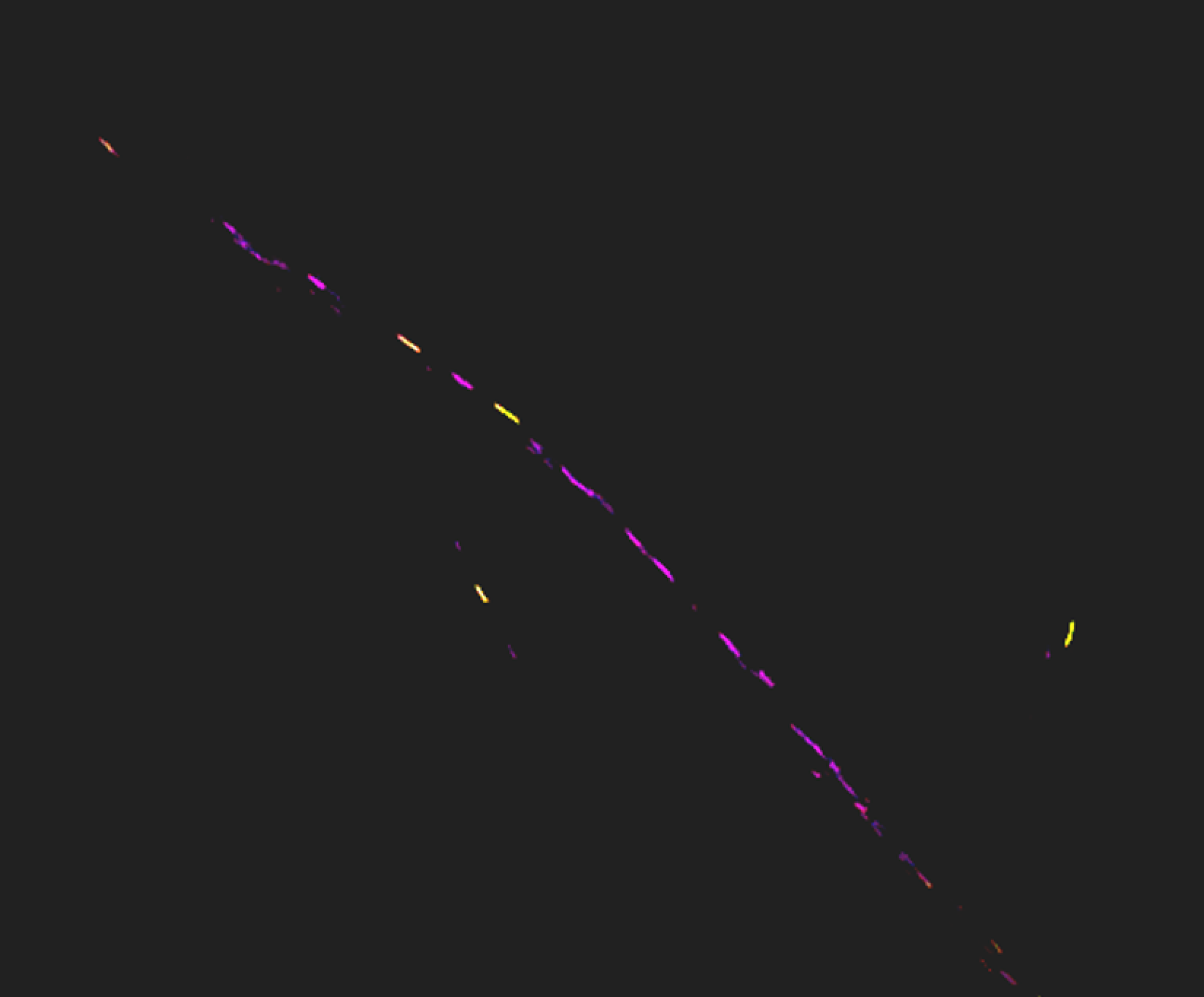} \\  
\end{tabular}
  \end{center}
  \vspace{-8pt}
  \caption{{\bf Top:} PlanetScope test imagery. {\bf Bottom:} Predicted mask from the PlanetScope segmentation model, showing moving cars in magenta and moving trucks in yellow.}

  \label{fig:preds0_ps}
\end{figure}

\begin{figure}
  \begin{center}
    \includegraphics[width=0.99\linewidth]{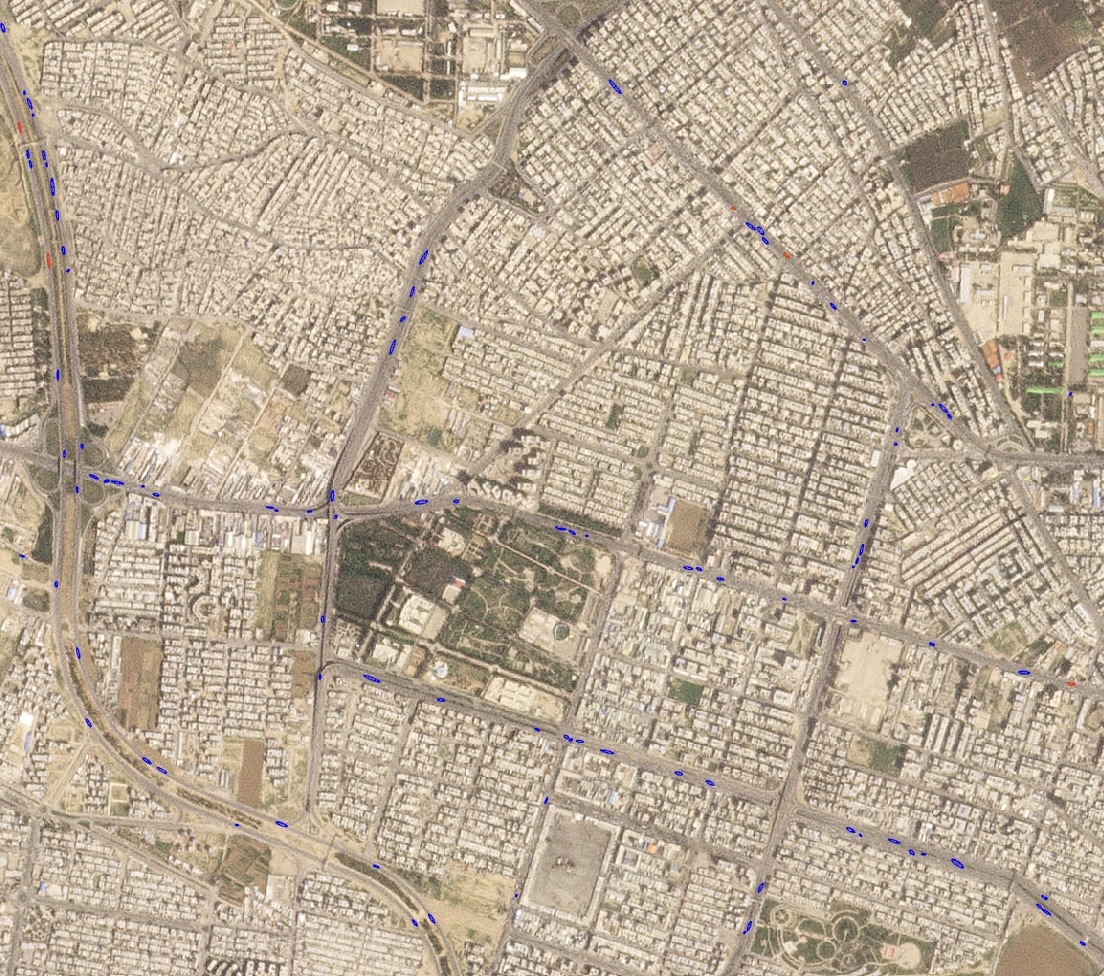}
  \end{center}
  \vspace{-0pt}
  \caption{Detected moving trucks (red ellipses) and cars (blue ellipses) overlaid on PlanetScope test imagery over Shiraz, Iran.}
  \label{fig:preds1_ps}
\end{figure}

While the 3m resolution of PlanetScope prevents us from detecting static vehicles as in SkySat, moving vehicle prediction masks are useful, see Figure \ref{fig:preds0_ps}.  Note that moving clouds cause a rainbow effect in Figure \ref{fig:preds0_ps}, though our segmentation algorithm is able to disentangle such effects from moving vehicles and this does not introduce false positives into the prediction mask.  Detecting moving cars is especially difficult in PlanetScope given that they occupy a mere $\sim 2$ pixels.  Ellipse detections are illustrated in Figure \ref{fig:preds1_ps}.  Performance is detailed in Table \ref{tab:ps_results}.  
F1 scores are low given the difficulty of the task and small training dataset, but counts are within $15\%$ of ground truth.  

\begin{table}
  \centering
  \begin{tabular}{@{}lcc@{}}
    \toprule
     {\bf Class} & {\bf F1} & {\bf Count Fraction} \\
    \midrule
    {Moving Car} & $0.28$ & $0.85$ \\
    {Moving Truck} & $0.49$  & $0.93$\\
    \midrule
    {Mean} & $0.39$  & $0.87$\\
    \bottomrule
  \end{tabular}
  \vspace{2pt}
  \caption{PlanetScope Performance.}
  \label{tab:ps_results}
  \end{table}

\subsection{Time Series Analysis}
\label{subsec:ts}

The frequent revisit of Planet's large satellite constellations enables the temporal dimension to be explored as well.  Precise time series analysis of SkySat imagery is possible when these tasked satellites pass over high interest areas.  

Here we focus on the lower spatial yet higher temporal resolution PlanetScope imagery.  With nearly daily revisit of the Earth's landmass, the PlanetScope constellation is ideally suited for exploring unexpected events, since a robust baseline of imagery exists even for all regions of the globe.  

\begin{figure}
  \begin{center}
    \includegraphics[width=0.98\linewidth]{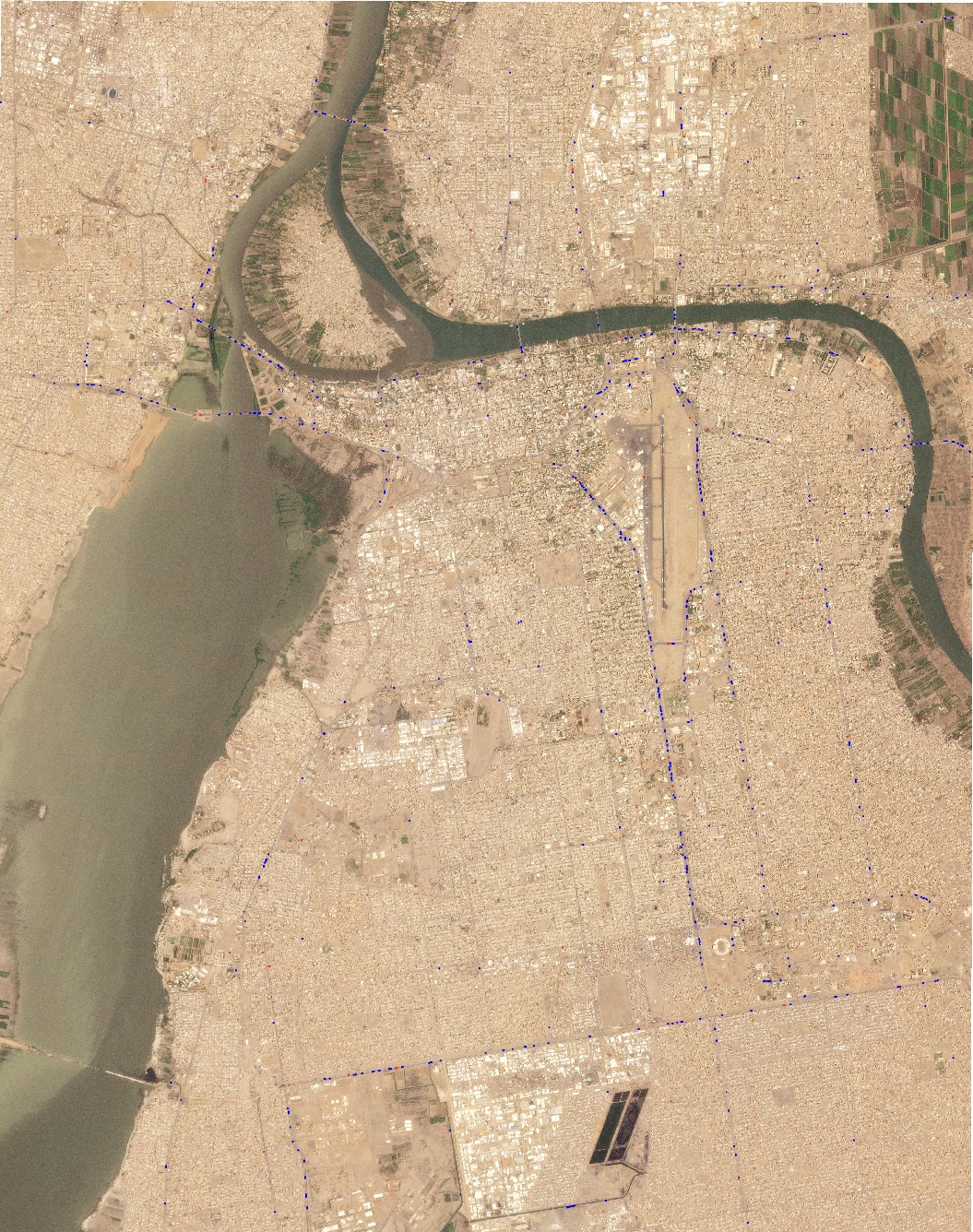}
  \end{center}
  \vspace{-0pt}
  \caption{Detected moving trucks (red ellipses) and cars (blue ellipses) overlaid on the Khartoum test region.}
  \label{fig:khartoum0}
\end{figure}

\begin{figure}
  \begin{center}
    \includegraphics[width=0.98\linewidth]{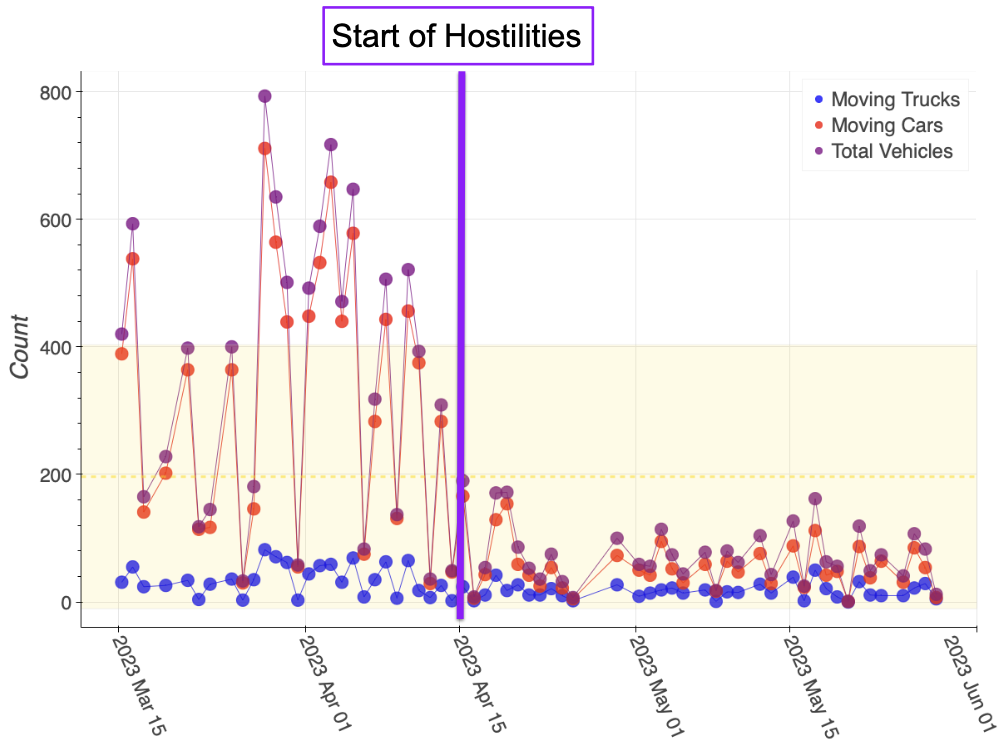}
  \end{center}
  \vspace{-0pt}
  \caption{Time series of moving vehicle detections in Khartoum in PlanetScope.}
  \label{fig:khartoum_ts}
\end{figure}

As an example, we explore a 280 km$^2$ region of central Khartoum, Sudan for 2.5 months in the Spring of 2023.  An example daily detection is shown in Figure \ref{fig:khartoum0}.  The time series of detected counts is shown in Figure \ref{fig:khartoum_ts}.  On 15 April, 2023 a civil war broke out in Sudan, with fighting concentrated in and around Khartoum.  Figure \ref{fig:khartoum_ts} indicates a notable change in volume occurs in mid-April 2023. 

\subsection{Vector Analysis}

\begin{figure}
  \begin{center}
  \begin{tabular}{cc}
 \includegraphics[width=0.48\linewidth]{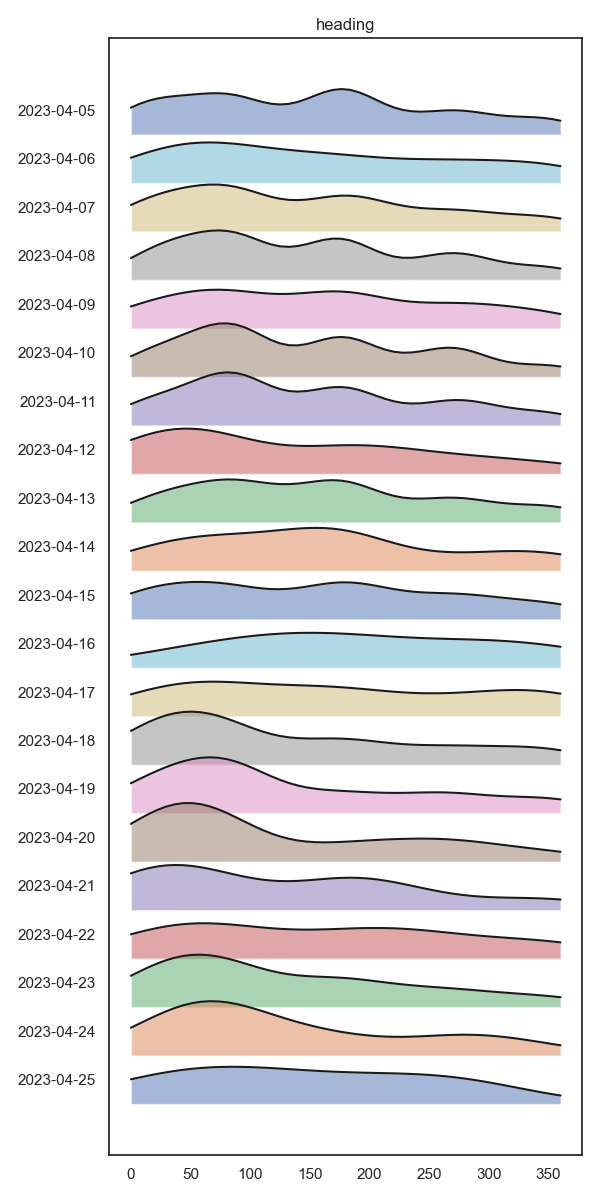} &  \includegraphics[width=0.5\linewidth]{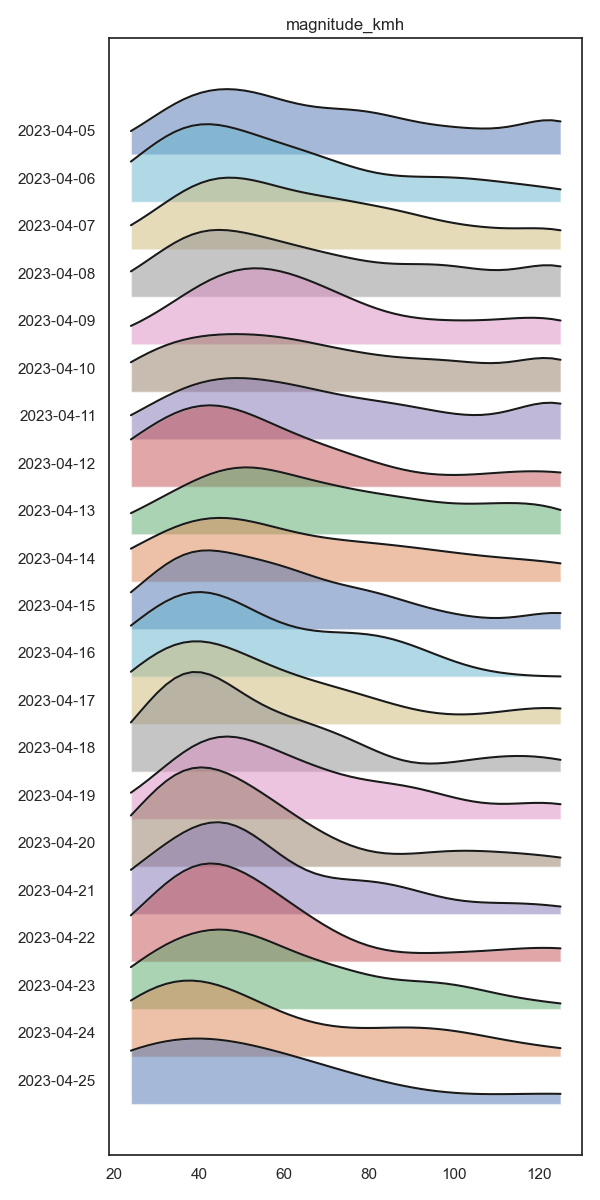} \\  
   [-0pt] 
 \includegraphics[width=0.48\linewidth]{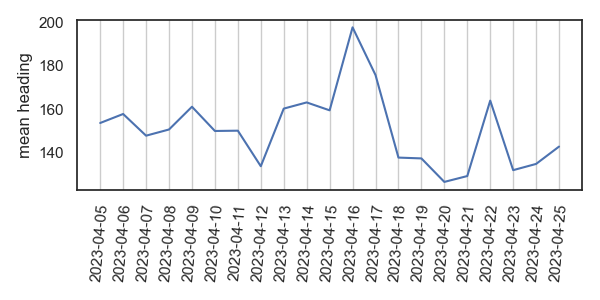} &  \includegraphics[width=0.5\linewidth]{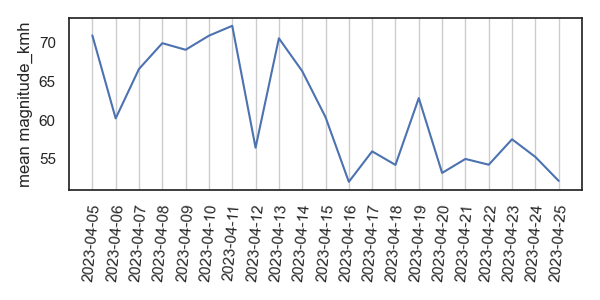} \\  
   [-0pt] 
\end{tabular}
  \end{center}
  \vspace{-0pt}
  \caption{ {\bf Upper Left:} Histogram of vehicle heading (degrees) in Khartoum. {\bf Upper Right:} Histogram of vehicle speed.  {\bf Lower Left:} Mean heading over time.  {\bf Lower Right:} Mean speed over time.}
  \label{fig:vec_ts}
  
   \vspace{30pt}

\end{figure}

In addition to the vehicle cardinality analysis performed in Section \ref{subsec:ts}, we can also explore vehicle movement vectors over time.  Figure \ref{fig:vec_ts} illustrates the magnitude and heading histograms in Khartoum for a three week period surrounding the commencement of the 2023 Sudanese civil war.  Inspection of this figure indicates that mean speed appears to reduce once the conflict begins.  The heading distribution also appears more uniform for the first few days of hostilities, hinting at a disruption of normal traffic patterns.

\newpage

\section{Conclusions}
\label{sec:conclusions}

In this paper we demonstrated the ability to localize ground vehicles in both high-resolution (0.5m SkySat) and medium-resolution (3m PlanetScope) Planet imagery.  Each constellation has its advantages, though each yields a distinctive motion-induced rainbow for non-static vehicles that can be algorithmically localized.  Furthermore, we are able to quantify the detected displacement vectors, thereby yielding a vector field of vehicle movement that provides estimates of both speed and heading.

In high-resolution SkySat imagery we are able to identify both static and moving cars with relatively high precision, achieve aggregate vehicle counts within $3\%$ of ground truth.  

In medium-resolution PlanetScope imagery we are not able to identify static vehicles, though are able to differentiate moving cars and moving trucks.  The daily global coverage of this constellation also enables robust time series analysis.  As a case study, we showed significant change in traffic volume and patterns following the commencement of hostilities in Khartoum, Sudan in April 2023.  Such time series analysis can easily be extended to any point on the globe given the ubiquitous coverage of this constellation.

We have only scratched the surface of what can be done with large area deep temporal stacks of vehicle vector fields.  Follow on work will further explore traffic pattern outliers.  We also hope to improve vehicle speed estimation by analyzing the raw grayscale Planet imagery (referred to as L0) taken directly from the spacecraft, rather than the processed and composited imagery analyzed here.  The composite imagery is far simpler to deal with and thus preferable for a proof of concept, but the individual L0 imaging frames could in theory be combined with a motion tracking algorithm to provide more precise speed estimates. 


\section*{Acknowledgements}

{\it Many thanks to Yoni Maryles for support and ideas.  Thanks to Adam Fritzler and Sara Bahloul for assistance with data collection specifics.}

\newpage

{\small
\bibliographystyle{ieee_fullname}
\bibliography{bib}
}

\end{document}